\documentclass[conference]{IEEEtran}  

\IEEEoverridecommandlockouts                              

\usepackage{amsmath,amsfonts}
\usepackage{algorithmic}
\usepackage{array}
\usepackage[abbreviations, british]{foreign}
\usepackage[caption=false,font=footnotesize,labelfont=rm,textfont=rm]{subfig}
\usepackage{textcomp}
\usepackage{stfloats}
\usepackage{url}
\usepackage{verbatim}
\usepackage{graphicx}
\usepackage{paralist}
\usepackage[percent]{overpic}
\usepackage[hidelinks]{hyperref}
\usepackage{csquotes}

\def\BibTeX{{\rm B\kern-.05em{\sc i\kern-.025em b}\kern-.08em
		T\kern-.1667em\lower.7ex\hbox{E}\kern-.125emX}}
\usepackage{balance}

\DeclareMathOperator*{\argmax}{arg\,max}

\usepackage{tikz}
\newcommand\copyrighttext{%
	\footnotesize \textcopyright 2024 IEEE. Personal use of this material is permitted.  Permission from IEEE must be obtained for all other uses, in any current or future media, including reprinting/republishing this material for advertising or promotional purposes, creating new collective works, for resale or redistribution to servers or lists, or reuse of any copyrighted component of this work in other works.}%
\newcommand\copyrightnotice{%
	\begin{tikzpicture}[remember picture,overlay]%
		\node[anchor=south,yshift=10pt] at (current page.south) {\fbox{\parbox{\dimexpr\textwidth-\fboxsep-\fboxrule\relax}{\copyrighttext}}};%
	\end{tikzpicture}%
}%

\usepackage{fancyhdr}
\fancypagestyle{firstpage}{%
\fancyhf{}%
\fancyhead[C]{Accepted for presentation at 2024 IEEE International Conference on Robotic Computing (IRC)}  
}

\author{Sascha Sucker, Michael Neubauer, Dominik Henrich
\thanks{S. Sucker, M. Neubauer, and D. Henrich are with the Chair of Applied Computer Science III (Robotics and Embedded Systems), University of Bayreuth, D-95440 Bayreuth, Germany;
	{\tt \{sascha.sucker | michael.neubauer | dominik.henrich\}@uni-bayreuth.de}}.%
}

\begin{document}

\title{Robot Tasks with Fuzzy Time Requirements\\from Natural Language Instructions}

\author{\IEEEauthorblockN{1\textsuperscript{st} Sascha Sucker}
\IEEEauthorblockA{\textit{Chair of Applied Computer Science III} \\
\textit{(Robotics and Embedded Systems)} \\
\textit{University of Bayreuth}\\
Bayreuth, Germany \\
sascha.sucker@uni-bayreuth.de}
\and
\IEEEauthorblockN{2\textsuperscript{nd} Michael Neubauer}
\IEEEauthorblockA{\textit{Chair of Applied Computer Science III} \\
\textit{(Robotics and Embedded Systems)} \\
\textit{University of Bayreuth}\\
Bayreuth, Germany \\
michael.neubauer@uni-bayreuth.de}
\and
\IEEEauthorblockN{3\textsuperscript{rd} Dominik Henrich}
\IEEEauthorblockA{\textit{Chair of Applied Computer Science III} \\
\textit{(Robotics and Embedded Systems)} \\
\textit{University of Bayreuth}\\
Bayreuth, Germany \\
dominik.henrich@uni-bayreuth.de}
}

\maketitle
\thispagestyle{firstpage}
\copyrightnotice%
\begin{abstract}
Natural language allows robot programming to be accessible to everyone.
However, the inherent fuzziness in natural language poses challenges for inflexible, traditional robot systems.
We focus on instructions with fuzzy time requirements (e.g., \enquote{start in a few minutes}).
Building on previous robotics research, we introduce fuzzy skills.
These define an execution by the robot with so-called satisfaction functions representing vague execution time requirements.
Such functions express a user's satisfaction over potential starting times for skill execution. 
When the robot handles multiple fuzzy skills, the satisfaction function provides a temporal tolerance window for execution, thus, enabling optimal scheduling based on satisfaction.
We generalized such functions based on individual user expectations with a user study.
The participants rated their satisfaction with an instruction's execution at various times.
Our investigations reveal that trapezoidal functions best approximate the users' satisfaction. 
Additionally, the results suggest that users are more lenient if the execution is specified further into the future.%

\end{abstract}


\section{Introduction}
\label{sec:introduction}
\begin{figure}[t]
	\centering
	\includegraphics[width=0.95\linewidth]{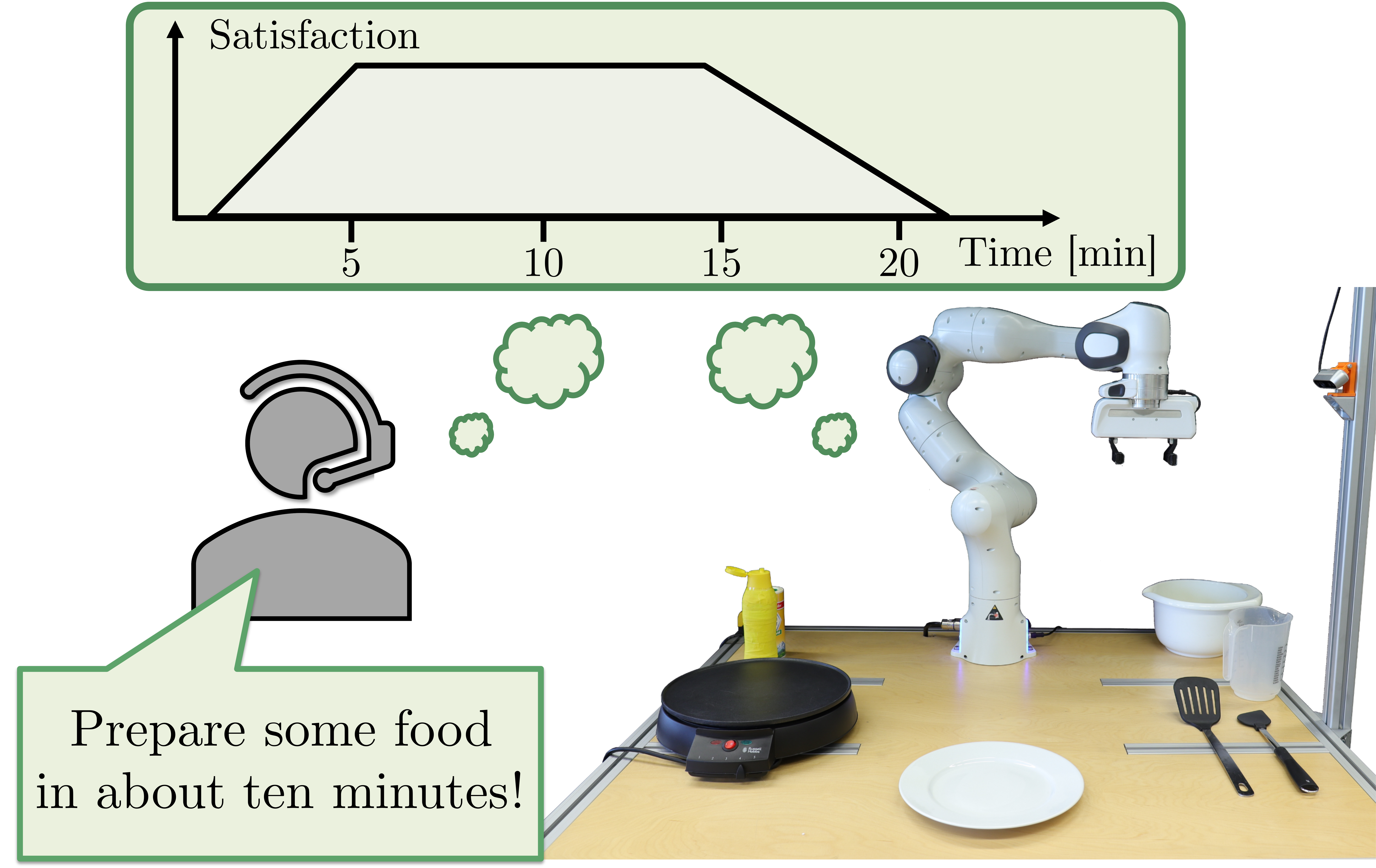}
	\caption{Natural language instructions are inherently fuzzy, requiring an interpretation within the context of the instruction and instructor. Here, the user's satisfaction varies over time based on the start of the task execution.}
	\label{img:motivation}
\end{figure}

Automating household tasks and production in small and medium enterprises is attracting considerable attention in robotics (e.g., \cite{Riedelbauch19, Dietz12}). 
Robot programming is still mostly relegated to specialized robotics experts, resulting in high costs and slow adaptation to new situations \cite{Dietz12}. 
One response to this is to increase the accessibility of robot programming \cite{Riedelbauch22, Hartwig23}, for example, with natural language \cite{Sucker23}.
Natural language contains inherent fuzziness that reduces the user's cognitive load. 
However, this contrasts with the rigid parameters (like concrete execution times) demanded by traditional robot systems \cite{Mavridis15}.
For example, the instruction ``Prepare some food in about ten minutes!'' (\autoref{img:motivation}) contains fuzziness regarding parameters (``some food'') and execution time (``about ten minutes'').
To handle fuzzy parameters, \textit{fuzzy-logic} \cite{Zadeh65} is commonly used to deduce exact parameters (e.g., weight in grams) for the operating robot system \cite{Klir95, Woelfel18}.
For time-dependent parameters, fuzzy-logic is extended to temporal fuzzy logic \cite{Dubois89}.
However, previous research focused hardly on the user's perception of fuzziness regarding execution time.
For instance in \autoref{img:motivation}, there is no immediate loss if the robot prepares the food five minutes earlier or later. 
Nevertheless, the user may be dissatisfied if the operation is not performed around the specified time, which depends on the instruction and the context. 
In this example, execution after 15 minutes leads to higher user satisfaction than after 20 minutes.

This paper investigates instructions with fuzzy time requirements.
An instruction describes a \textit{fuzzy skill} that encodes the manipulation of an object by the robot (in continuation of \cite{Pedersen16, Fichtner23}).
If the user instructs only one fuzzy skill, it can be executed at maximum satisfaction by default.
However, suppose the user instructs several fuzzy skills combined into one superordinate plan, the \textit{fuzzy task}. 
In that case, the task execution may require compromises, i.e., the robot must perform some skills at a suboptimal time. 
For example, if the user issues another command that should also start in ten minutes (\autoref{img:motivation}), rigid robot systems lack the knowledge of which operation to prefer.
The satisfaction function provides this required knowledge -- enabling the scheduling to maximize overall satisfaction.
Precisely identifying the expectations of an individual user before execution is challenging.
In addition to the given instruction itself, other aspects could also influence satisfaction, e.g., the context of the scenario, the user's previous experience, or the expected abilities of the actor.
Another challenge is that several users with (partly) divergent expectations may instruct the robot system.
One response to this could be the creation of user profiles.
However, this does not account for frequent user changes or public scenarios with previously unknown users.
Hence, we aim to provide general statements about deriving the satisfaction functions from the instructions and context.


We present two central contributions: 
\begin{inparaenum}[(i)]
	\item We formalize fuzzy tasks, their inference from language, and the scheduling (\autoref{sec:methodology:tafs} and \autoref{sec:methodology:handling}).
	\item We deduce an overall satisfaction function from individual user satisfaction (\autoref{sec:methodology:identifying}).
\end{inparaenum}
On this basis, we examine fuzzy time requirements regarding their modeling, the difference between human and robot actors, and the influence of time until the required execution starts (\autoref{sec:evaluation}).
For this, we exploit subjective satisfaction data gathered with an online user study. 


\section{Related Work}


The resolution of uncertainties is already examined within the contexts of robotics \cite{Carinena04, Abou12}, control systems \cite{Li09, Martos08}, and scheduling system \cite{Dubois03, Slany94, Liu21}. 
These approaches use fuzzy sets defined by membership functions to determine exact output values from vague input statements. 
Within natural language processing, fuzzy sets can also represent natural language components to quantify fuzziness \cite{Hersh76, Woelfel18}. 
For example, language-based decision-making systems \cite{Omoregbe20} or information retrieval systems \cite{Bolc85, Tamrakar14} utilize this quantification.
However, these approaches use expert knowledge to set the membership function of a corresponding fuzzy set, which may not always be available. 
Incoming data can parameterize membership functions \cite{Cerrada05, Zhu10}, but this was not investigated with natural language.

Similarly, no fuzzy sets examined temporal information for instruction/skill sorting or scheduling. 
Usually, in natural language, the order of instructions does not always adhere to logical-temporal relationships \cite{Pane00, Lieberman06}.
An approach that achieves a chronological sorting of natural language statements uses logical-temporal relationships between expressions \cite{Landhausser14}. 
Here, temporal keywords (e.g., 'before', 'after', and 'while') indicate the order of the expressions, thus, forming a timeline.
Additionally, the context of instructions may be interpreted to find a sensible execution order (e.g., 'cleaning' should be done after 'drilling') \cite{Liu18}.
This context is especially required if the expressions lack temporal keywords.
Other approaches tackle sorting natural language statements with linear temporal logic (LTL) \cite{Liu22, Cosler23}.
The transformation of natural language to logical expressions in LTL is challenging.
This problem is solved using large language models like BERT \cite{Devlin18} or GPT-4 \cite{Achiam23}. 
Large language models are pre-trained neural networks that return LTL expressions when prompted.
These expressions are then used to schedule the actions within the text. 
The approaches using temporal operators \cite{Landhausser14} or LTL (e.g., \cite{Liu22, Cosler23}) do not provide time values required for scheduling with fuzzy time requirements but rather dependencies between expressions.
Thus, a formalization of tasks with such fuzzy time requirements is missing.
Studies investigating natural language as a medium for programming (e.g., \cite{Pane00, Lieberman06}) have yet to examine the impact of fuzzy time requirements on satisfaction regarding execution time.
Our work addresses these gaps.

\section{Methodology}
\begin{figure}[t]
	\centering
	\begin{overpic}[width=\linewidth]{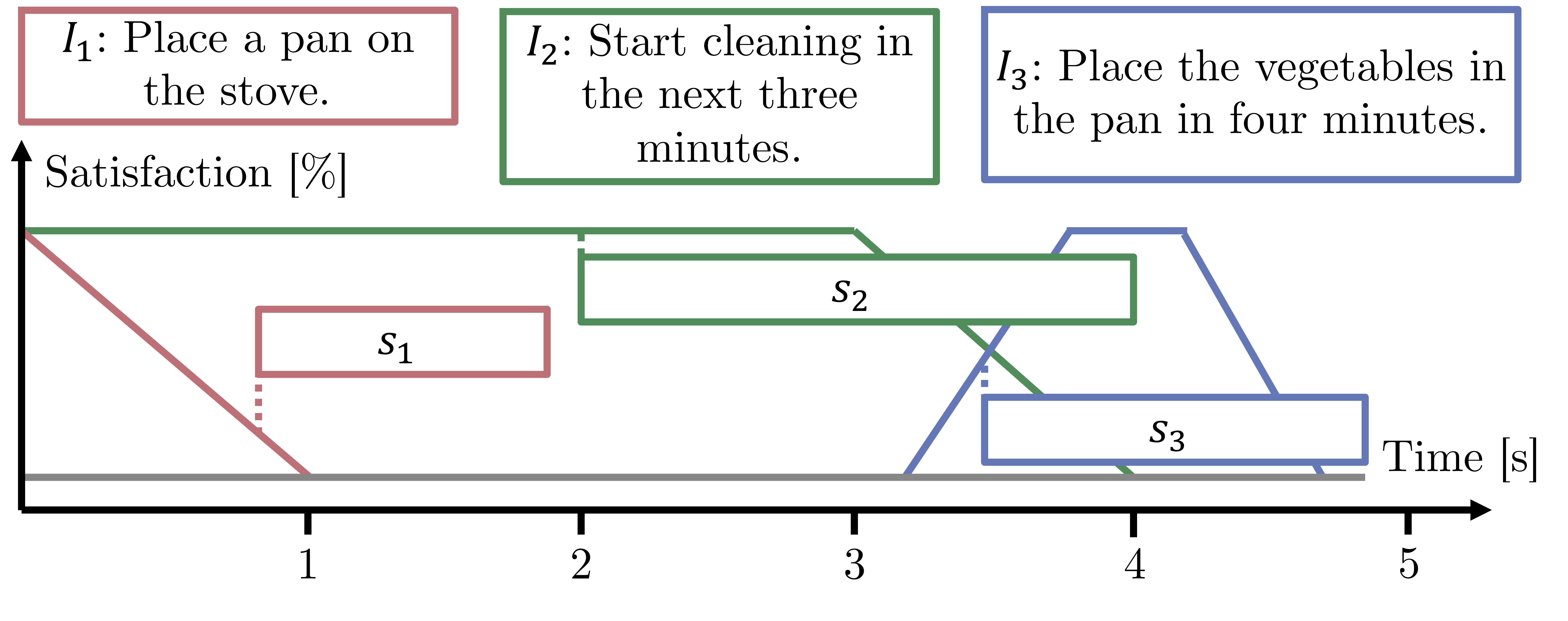}
	\end{overpic}
	\caption{
		The actor must cope with fuzzy time specifications for each instruction $I_i$ (issued at $0$ s). We consider the user satisfaction ($\psi_i$) when the specific skills ($s_i$) are started, indicated by the boxes $s_1$ to $s_3$. Non-optimized scheduling may lead to low satisfaction ($s_1$, $s_3$) and overlaps. } 
	\label{img:membership_functions}
\end{figure}

This paper focuses on assessing the perception of fuzzy time requirements in natural language instructions.
For this purpose, we model such fuzzy time requirements with functions representing the user's satisfaction regarding the execution over time (\autoref{sec:methodology:tafs}).
We interpret satisfaction functions from natural language instructions and their context (\autoref{img:membership_functions}).
Multiple satisfaction functions within a task are used for scheduling with fuzzy deadlines.
We define the corresponding optimization problem with additional restrictions, including overlap.
In \autoref{sec:methodology:handling}, we formalize this problem and outline solutions.
Satisfaction functions describe the subjective user expectations concerning the execution times of the task.
We derive general satisfaction functions from these subjective satisfaction functions (\autoref{sec:methodology:identifying}).
These methodologies serve as the basis for our evaluation (\autoref{sec:evaluation}).

\subsection{Tasks with Fuzzy Time Requirements}
\label{sec:methodology:tafs}
A \textit{task} is the model describing the manipulation of objects.
Our tasks comprise skills, each describing the manipulation of objects within the given scene.
\textit{Skills} serve as mid-level representations of robot operations \cite{Pedersen16} abstracting from motion primitives (explicit robot movements) but providing more detail than the overarching tasks.
They are the most specific operation representation that manipulates objects independently of any particular hardware setup.
We distinguish between specific and fuzzy skills.
\textit{Specific skills} are defined following the commonly used skills with no ambiguities (as in \cite{Pedersen16, Paxton17}).
They consist of motion primitives that are entirely and precisely parameterized.
Following \cite{Pedersen16}, specific skills can be directly executed, thereby manipulating objects based on the skill's composition and the robot's capabilities.
In addition to the common definition in \cite{Pedersen16}, we focus on task time requirements.
Thus, we require our skills to contain such time requirements -- leading to the specific skill tuple 
\begin{align}
	s_i=(t^s_i, t^d_i, \Lambda_i).
	\label{eq:specific_skill}
\end{align}
It contains the explicit start time and duration $t^s_i, t^d_i \in \mathbb{R}_+$ and additional, specific parameters $\Lambda_i$. 
The origin of the timeline for both the positive real-valued times $t_s$ and $t_d$ is when the instruction is issued.
The parameter set $\Lambda_i$ represents every non-time-related parameter (e.g., the object to be manipulated or its goal state).
This is purposely kept abstract to transfer our concepts to various robot systems.

\textit{Fuzzy skills} represent vague operation specifications (\autoref{img:motivation}).
This vagueness constitutes fuzzy object descriptions, parameters, and time requirements.
The concrete mapping from fuzzy object descriptions to physical objects is discussed in our prior works \cite{Riedelbauch22, Sucker23}. 
Fuzziness in parameters is frequently tackled with (temporal) fuzzy logic (e.g., in \cite{Woelfel19}).
Consequently, we primarily focus on fuzzy execution time requirements.
We define the parameters for fuzzy skills
\begin{align}
	\Tilde{s_i}=(\psi_i, t^d_i, \Theta_i, \Tilde{\Lambda_i}),
	\label{eq:fuzzy_skill}
\end{align}
as a tuple of the satisfaction function $\psi(t)$ and the duration $t^d_i$ (\autoref{eq:specific_skill}) as well as a mix of fuzzy $\Theta_i$ and specific parameters $\Tilde{\Lambda_i}$.
The \textit{satisfaction function} 
\begin{align}
	\psi: \mathbb{R}_+ \rightarrow \left[ 0, 1 \right]
	\label{eq:satisfaction_function}
\end{align}
maps positive real-valued time ($\mathbb{R}_+$) to a satisfaction value between $0$ and $1$, indicating a spectrum between no to complete satisfaction (inspired by real-time requirements \cite{Kopetz22}).
To streamline scheduling, we define the input time of $\psi$ as the start time of the specific skill.
This requirement does not imply that the specified time within instructions always refers to the start time.
The instruction may contain \textit{indicator verbs} that relate to the start, end, or weighted time over the execution duration (e.g., 'start' or 'finish').
The indicator verb 'start' ($I_2$ in \autoref{img:membership_functions}) references the start time of the operation, while 'finish' would indicate an end time. 
This indicator verb may be omitted ($I_1$ and $I_3$ in \autoref{img:membership_functions}), requiring an inference by context.
When converting the instruction into a fuzzy skill, the indicator verb should be considered.
However, indicator verbs are beyond the scope of this paper.
Thus, we assume that all instructions in this conversion (\autoref{sec:methodology:inferring}) and the evaluation (\autoref{sec:evaluation}) refer to the start time.
A \textit{fuzzy task} is a set of fuzzy skills (\autoref{img:membership_functions}).
Fuzzy tasks are unordered as the execution times rely on the skills' satisfaction functions.

\subsection{Handling Tasks with Fuzzy Time Requirements} 
\label{sec:methodology:handling}
Robot systems must cope with uncertainties in natural language instructions.
This involves deriving the fuzzy skills from instructions and converting them into specific skills for execution.
For this purpose, an instruction $I$ is converted to a fuzzy skill $\Tilde{s}$.
We mainly follow our previous work \cite{Sucker23} for this mapping.
Nevertheless, we consider fuzzy temporal constraints as part of the task description.
To this end, suitable parts-of-speech are extracted from the instructions and interpreted according to the context, yielding fuzzy skills (\autoref{sec:methodology:inferring}).
To convert them into specific skills, the start time and explicit parameters are derived from a set $T=\{\Tilde{s_1}, ..., \Tilde{s_n}\}$ of $n$ fuzzy skills.
The start time is determined by maximizing the satisfaction of every fuzzy skill without overlapping executions.
For this, the task must be considered as a whole -- leading to a scheduling problem, presented in \autoref{sec:methodology:scheduling}.
Three possible solutions are discussed: an exhaustive exploration of the problem space, hill climbing, and simulated annealing.

\subsubsection{Inferring Fuzzy Time Requirements from Language}
\label{sec:methodology:inferring}
\begin{figure}
	\centering
	\includegraphics[width=\linewidth]{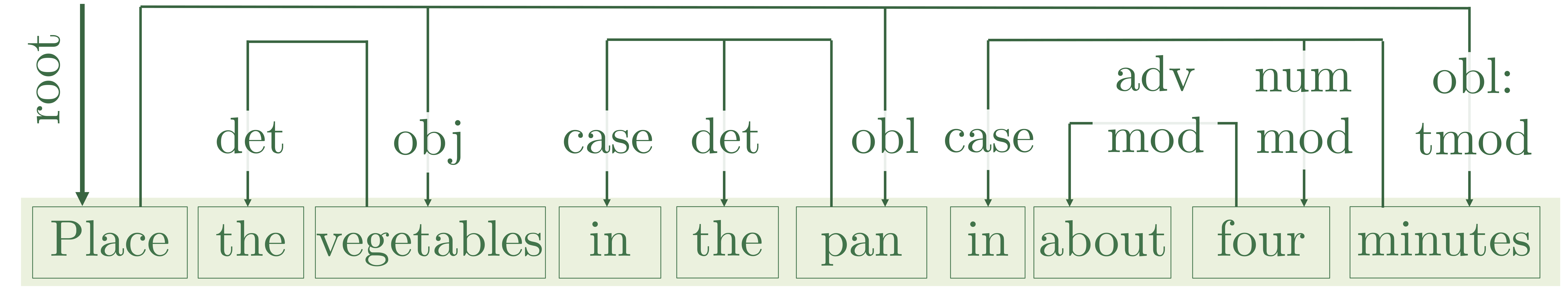}
	\caption{The dependency tree encodes grammatical connections between words in a sentence. These dependencies include oblique temporal modifiers (obl:tmod), numbers (nummod), adverbs (advmod), and prepositions (case).}
	\label{img:dependency_tree}
\end{figure}

We extend our previous work \cite{Sucker23} by identifying and interpreting temporal specifications in natural language instructions.
Current approaches utilize Large Language Models (LLMs) for natural language programming \cite{Singh23, Vemprala24} in particular, as they deliver good results in previously unknown environments.
However, these approaches still show some problems, such as transparency, explainability, and consistency.
Since the language interpretation only is secondary to this paper and the investigation of LLM robot programming is still early, we opted for an established approach:
We perform a \textit{syntactic analysis} of the instruction, i.e., a \textit{part-of-speech tagging} \cite{Jurafsky23} together with the construction of a \textit{dependency tree} \cite{Marneffe14}.
The dependency tree describes how the words in a sentence are interconnected syntactically.
Prior work \cite{Landhausser14, Ferro01} already focuses on sorting instructions if they directly relate to each other.
Thus, we concentrate on instructions that refer to points in time (``in about three minutes''), i.e., to identify temporal specifications that refer to the instructed operation (verb).
Accordingly, we search for an oblique temporal modifier dependent on the root verb in the dependency tree. 
In the instruction in \autoref{img:dependency_tree}, the temporal modifier 'minutes' is referenced from the root verb 'Place'.
Further specifiers are searched recursively, e.g., number modifiers ('four'), fuzziness modifiers ('about'), and prepositions ('in').

The satisfaction $\psi$ must be derived from these parameters.
One possibility is a \textit{direct lookup}:
A satisfaction function is stored for each possible modifier combination, which is looked up in the corresponding instruction.
However, a separate function would be required for every modifier combination (e.g., for every possible number modifier).
Instead, there should only be a few fundamental functions that are adapted depending on the modifier, i.e., \textit{lookup and adapt}.
Here, for example, the function could be scaled due to fuzziness modifiers like 'approximately' ($\psi(m_p \cdot t)$ with $m_p \neq 0$) or shifted due to the number modifier ($\psi(t + m_n)$ with $m_n \geq 0$).
The width of the satisfaction may also change due to the number modifier.
However, such lookup methods do not take the context of the instruction into account. 
For example, the pan temperature can significantly influence when the instruction $I_3$ should be executed (cf. \autoref{img:motivation}).
Temporal Fuzzy-Logic can be used for this purpose.
A system implementing Temporal Fuzzy-Logic holds a set of predefined rules. 
These rules determine the satisfaction function model and its parameters based on a combined set of context information. 
For example, in the instruction of \autoref{img:dependency_tree}, one may utilize the pan's heat to find a suitable execution time.
Context information may be especially valuable if no explicit time is specified (e.g., 'soon').



\begin{figure*}[!t]
	\centering
	\subfloat[Satisfaction functions of ten users.]{
		\includegraphics[width=0.29\linewidth]{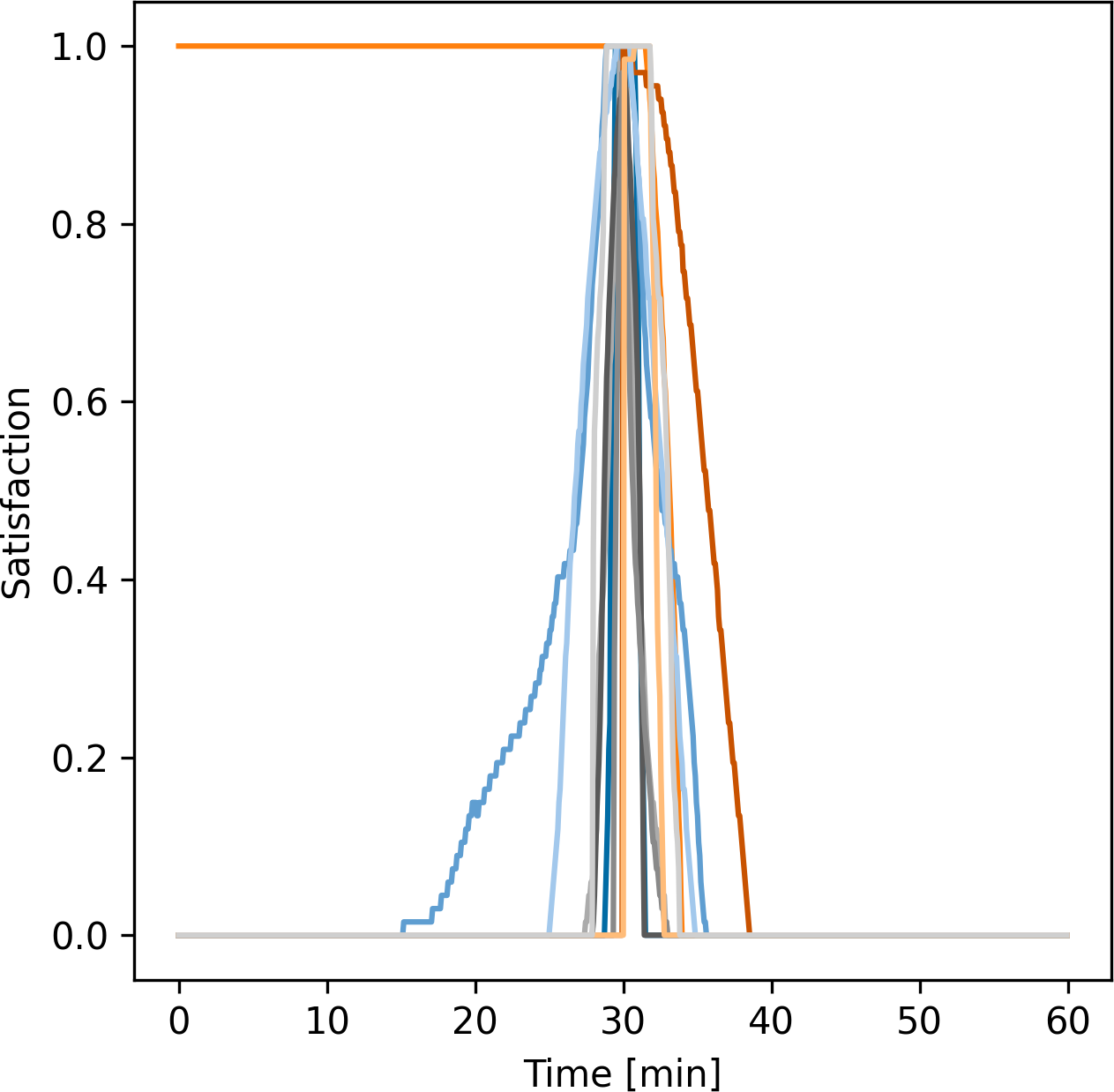}%
		\label{img:satisfaction_functions}
	}
	\hfill
	\subfloat[Histogram of satisfaction functions.]{
		\includegraphics[width=0.3525\linewidth]{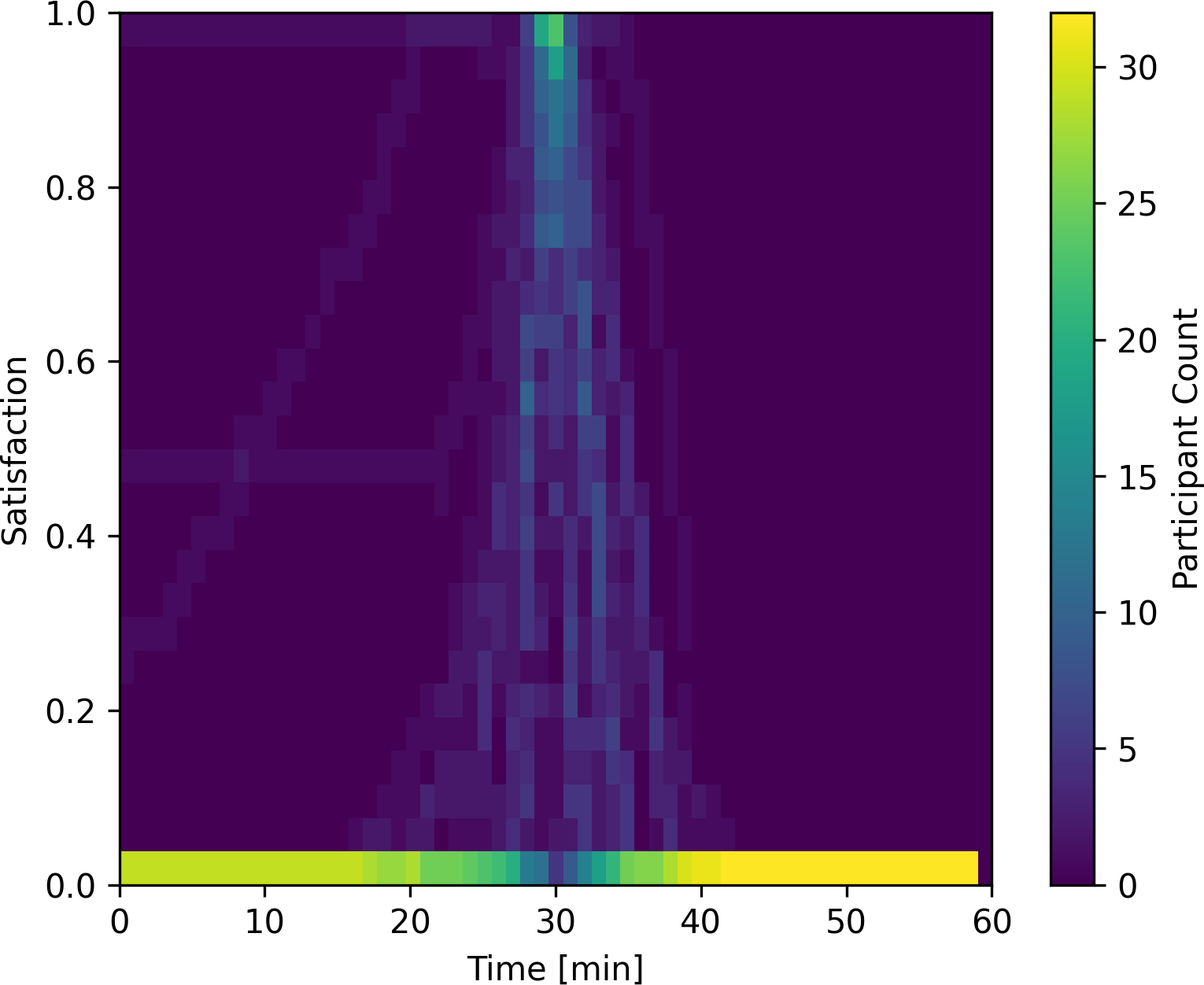}%
		\label{img:histogram}
	}
	\hfill	
	\subfloat[Distribution plot of satisfaction functions.]{
		\includegraphics[width=0.29\linewidth]{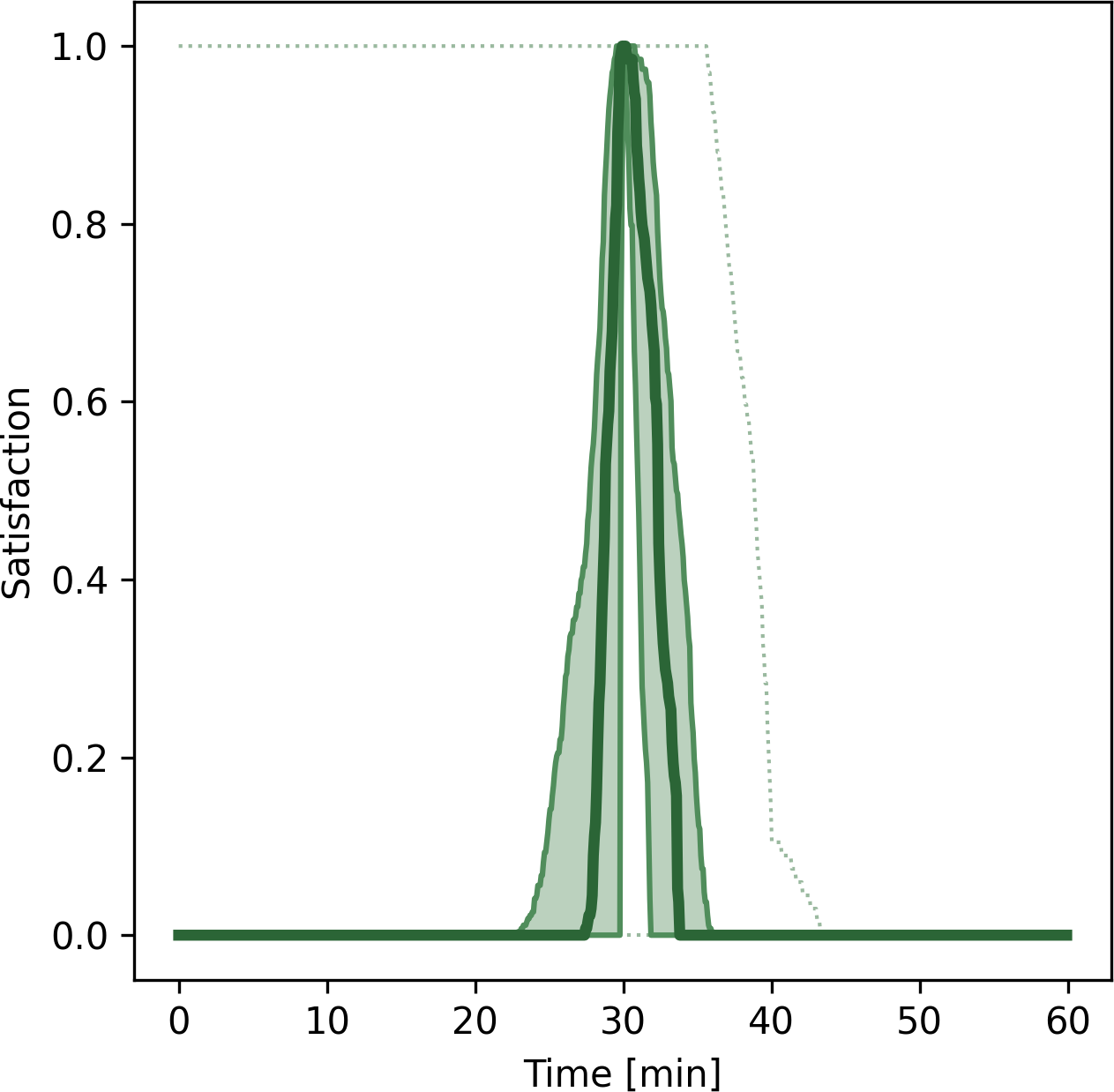}%
		\label{img:boxplot}
	}
	\caption{Users' expectations regarding executing one instruction may vary. Given the instruction ``The assignment should start in 30 minutes!'', our study participants drew their satisfaction functions. In (a), only ten functions are shown for a better overview.
		The distribution across all participants is displayed using a histogram (b) or distribution plot (c). In c), the area between the 25- and 75-quantiles is shaded; the minimum and maximum values are dotted.  }
	\label{fig_sim}
\end{figure*}

\subsubsection{Scheduling with Fuzzy Time Requirements}
\label{sec:methodology:scheduling}
A fuzzy task is scheduled to maximize the user's satisfaction.
We assume that only one agent handles the skills in the task.
For this reason, the skill executions may not overlap in the resulting schedule.
Given a fuzzy task $T=\{\Tilde{s_1}, ..., \Tilde{s_n}\}$ as a set of $n$ fuzzy skills (\autoref{eq:fuzzy_skill}), we search for the optimal start time vector $\vec{t^*} \in \mathbb{R}^n_+$. 
It describes the start time of each fuzzy skill $\Tilde{s_i}$ that optimizes satisfaction without overlap.
Regarding satisfaction optimization, we search for 
\begin{align}
	\vec{t^*}=\argmax_{(t_1, ..., t_n)\in\mathbb{R}^n_+}\,\prod^n_{i=1} \psi_i(t_i).
	\label{eq:satisfaction_optimization}
\end{align}
To avoid overlaps, no two skill execution time intervals may intersect.
To test this, we sort the elements of $\vec{t^*}$ using a sort function $\pi: \{1,...,n\} \rightarrow \{1,...,n\}$ defined by $t_{\pi(i)}\leq t_{\pi(i+1)}, \forall i \in \{1,...,n-1\}$.
Given the duration $t^d_{\pi(i)}$, the overlap avoidance restricts the solution space by requiring
\begin{align}
	t_{\pi(i)}+t^d_{\pi(i)} \leq t_{\pi(i+1)} \enspace \forall i\in\{1, ..., n-1\}.
	\label{eq:no_overlap}
\end{align}
After optimization (\autoref{eq:satisfaction_optimization}) with overlap avoidance (\autoref{eq:no_overlap}), the start time of the specific skill $s_i$ is $t_i \in \vec{t^*}$.

Conceptually, we search for the optimization of all $\psi$ by linking them with the logical AND, leading them to be multiplied.
This results in the required decisive satisfaction loss if one or a few start times are hardly accepted.
However, the optimization fails if at least one satisfaction results in $0$, e.g., due to the overlap avoidance.
Thus, the overall satisfaction is $0$ independent of the other start times. 
Accordingly, start times for other skills could be chosen arbitrarily, not influencing overall satisfaction.
Arguably, the task would be invalid if every arrangement of skills leads to one satisfaction of $0$:
The execution of this skill would not be accepted.
However, if optimization of the remaining skills is still desired, the output of the satisfaction functions $\psi$ can be adjusted.
One practical solution could be to set a minimum threshold for $\psi_i(t) > 0$.

Satisfaction can be adapted for various scheduling applications, e.g., by associating it with general execution utility. 
In contrast, the overlap avoidance may be relaxed if $m$ skills can be executed in parallel, e.g., by several robots.
Accordingly, the constraint would be changed to a maximum of $m$ overlapping skills at one single time step.
Based on the application, other aspects may be considered for scheduling, including the total execution time, the idle time, the intrusiveness to the instructor, or the energy consumption.


To solve the optimization, a combinatorial approach is only feasible to a limited extent:
The search space $\mathbb{R}^n_+$ is continuous, resulting in infinite possible discrete solutions.
One naive approach is to sample $\mathbb{R}^n_+$.
For this purpose, the considered time interval $J_s$ is restricted and sampled at a frequency $u$.
This results in $k\in \mathbb{N}$ \textit{time steps} $k = \lfloor u \cdot |J_s| \rfloor$,
where $|J_s|$ denotes the length of the time interval $J_s$.
All sampled start times for $\vec{t^*}$ are tested for each skill.
This leads to an exponential runtime of $O(k^n)$, rendering it feasible only for small search spaces.
Consequently, the solution must be approximated even for moderately large problem spaces.

For this, a possible approach is the Hill Climbing algorithm \cite{Russell10}. 
It tries to maximize the schedule's satisfaction by shifting the skills' start times. 
Such a greedy approach, however, runs the risk of returning a sub-optimal solution by converging within a local optimum. 
To solve this issue, Simulated Annealing may be utilized \cite{Russell10}. 
This algorithm mostly follows Hill Climbing but allows for a reduction in satisfaction based on probability.
With this, the algorithm may escape local optima but requires more iterations.  
Another approach is to utilize genetic algorithms for optimizing scheduling by mimicking evolutionary processes of selection and reproduction \cite{Russell10}.





\subsection{Estimating Satisfaction from User Data}
\label{sec:methodology:identifying}
The satisfaction $\psi(t)$ of a fuzzy skill depends, by definition, on the subjective expectations of the user.
However, we aim to determine general statements about deriving the satisfaction functions from the instructions and context.
For this purpose, the expectations of multiple users must be analyzed, e.g., through a user study (cf. \autoref{sec:evaluation}).
Given one specific instruction $I$, a user $u_i$ states their satisfaction within $\psi_i$ (for example, by drawing). 
The satisfaction functions of multiple users are combined to the \textit{subjective satisfaction functions} as a set $\Psi_I=\{\psi_1, \psi_2, ...\}$ of satisfaction functions (\autoref{img:satisfaction_functions}).
If a sufficient number of users provide their satisfaction, the satisfaction at each time step forms a probability function.
Thus, the subjective satisfaction functions $\Psi_I$ comprise three dimensions: 
satisfaction (\cf \autoref{img:membership_functions}), time, and distribution of the individual satisfaction functions.
In \autoref{img:histogram}, we represent this using a three-dimensional graph whose distribution is approximated by a histogram.
From $\Psi_I$ we determine the satisfaction $\psi(t)$.
Furthermore, by analyzing $\Psi_I$, we can identify aspects that influence satisfaction functions.


\subsubsection{Determining Characteristic Values}
\label{sec:methodology:discrete_satisfaction}
We analyze the subjective satisfaction function by viewing each time step as a probability function and determining characteristic values.
For example, we calculate arithmetic mean $\overline{\Psi_{I}}(t)$ point-wise for each time step $t$ with
\begin{align}
	\overline{\Psi_{I}}(t) = \frac{1}{|\Psi_I|} \sum_{\psi \in \Psi_I} \psi(t).
	\label{eq:arithmetic}
\end{align}
This is then examined over each time step, allowing us to draw general conclusions.
The generalization quality for $\Psi_I$ to one $\psi$ is expressed with quality metrics, e.g., deviation.
A high deviation indicates conflicting participant expectations, which may require dividing the users into subgroups.
With this, the satisfaction function could better represent user expectations.
The overall satisfaction function may be calculated considering the mode, arithmetic mean, and median (each evaluated point-wise).
Although the mode represents the most frequent satisfaction, it may not represent the central tendency of the distribution (e.g., in the case of several maxima or asymmetry).
The mean value indicates the central tendency but is susceptible to outliers compared to the median.
Thus, we primarily focus on the median during our evaluation.
In addition, quantiles help to visualize $\Psi_I$, for example, in a distribution plot (\autoref{img:boxplot}).
Other values, such as skew or kurtosis, could also be used to analyze the function.

Such a satisfaction function is discretized to the time steps of the subjective satisfaction functions, for example, when the satisfaction for individual time steps is recorded in a user study.
If the discretization between this satisfaction function and the scheduling does not match, an interpolation between the values must be performed, requiring additional computational effort.
This approach also forces the satisfaction values to be saved for each time step, requiring extensive memory.
Reusing (and thus referencing) such satisfaction functions is challenging, as instructions can contain arbitrary time specifications with variable specificity.
For a task plan with $n$ fuzzy skills and discretization in $k$ time steps, a memory overhead of $O(n \cdot k)$ arises. 
This can be addressed by further approximating the satisfaction.


\subsubsection{Fitting Continuous Satisfaction Functions} 
\label{sec:methodology:continuous_satisfaction}
For this purpose, we find a continuous best-fitting function for the satisfaction function $\psi$.
Accordingly, no interpolation is required for scheduling. 
Instead, the function is evaluated at the respective time values.
Depending on the complexity of the continuous function, the calculation effort can be similar to interpolation.
Nevertheless, only a fixed number of parameters must be held to describe the function, regardless of the time steps $k$.
Various models and methods can approximate such a function.
For example, the function may be approximated utilizing polynomials \cite{James13}. 
However, a better approximation requires a higher polynomial degree, leading to extensive memory and calculation requirements.

Alternatively, predetermined functions could be fitted to the data, for example, with the \textit{Trust Region Reflective} algorithm \cite{Branch99}.
With this algorithm, non-linear functions can also be fitted to the data (e.g., bell curves or trapezoidal functions).
The algorithm iteratively adjusts the search region within which it approximates the objective function with the model to be fitted. 
However, this algorithm may get stuck in local minima, so choosing initial parameters is particularly important.
In general, approximation reduces the accuracy of the satisfaction function.
Nevertheless, the $k$ time steps of the satisfaction function can be reduced to a few parameters.
Such an approximation can be especially beneficial if human expectations and specific models coincide, resulting in a negligible error.


\begin{figure}[t]
	\centering
	\includegraphics[width=\linewidth]{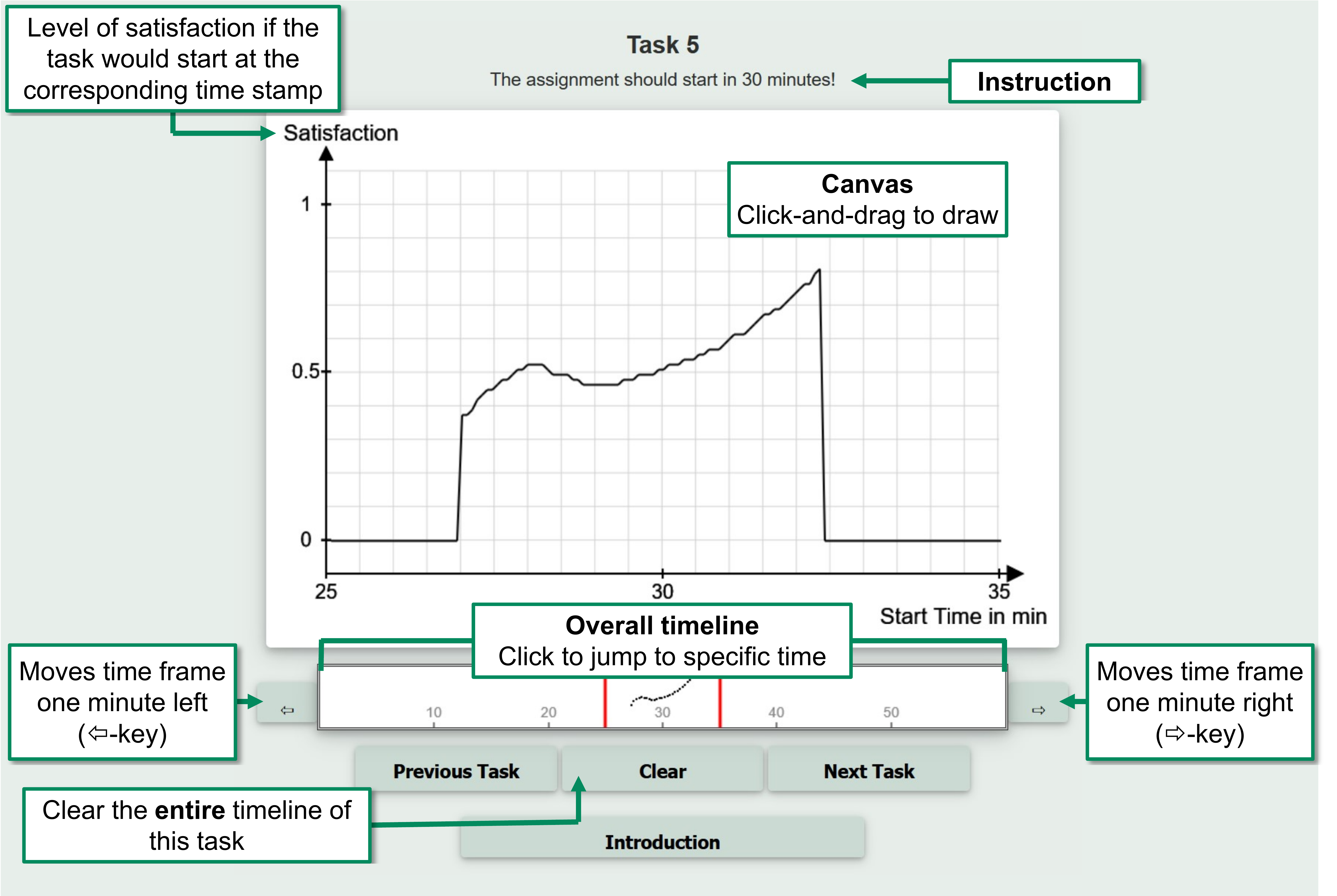}
	\caption{
		Participants of our online user study should draw their satisfaction given instructions with fuzzy time requirements if the task started at the time stamp. 
		Before carrying out the first task, this tutorial page was shown to the participants --
		highlighting the core functionalities of the interface. } 
	\label{img:tutorial_study}
\end{figure}

\begin{figure*}[!t]
	\centering
	\subfloat[``in ten minutes'']{
		\includegraphics[width=0.31\linewidth]{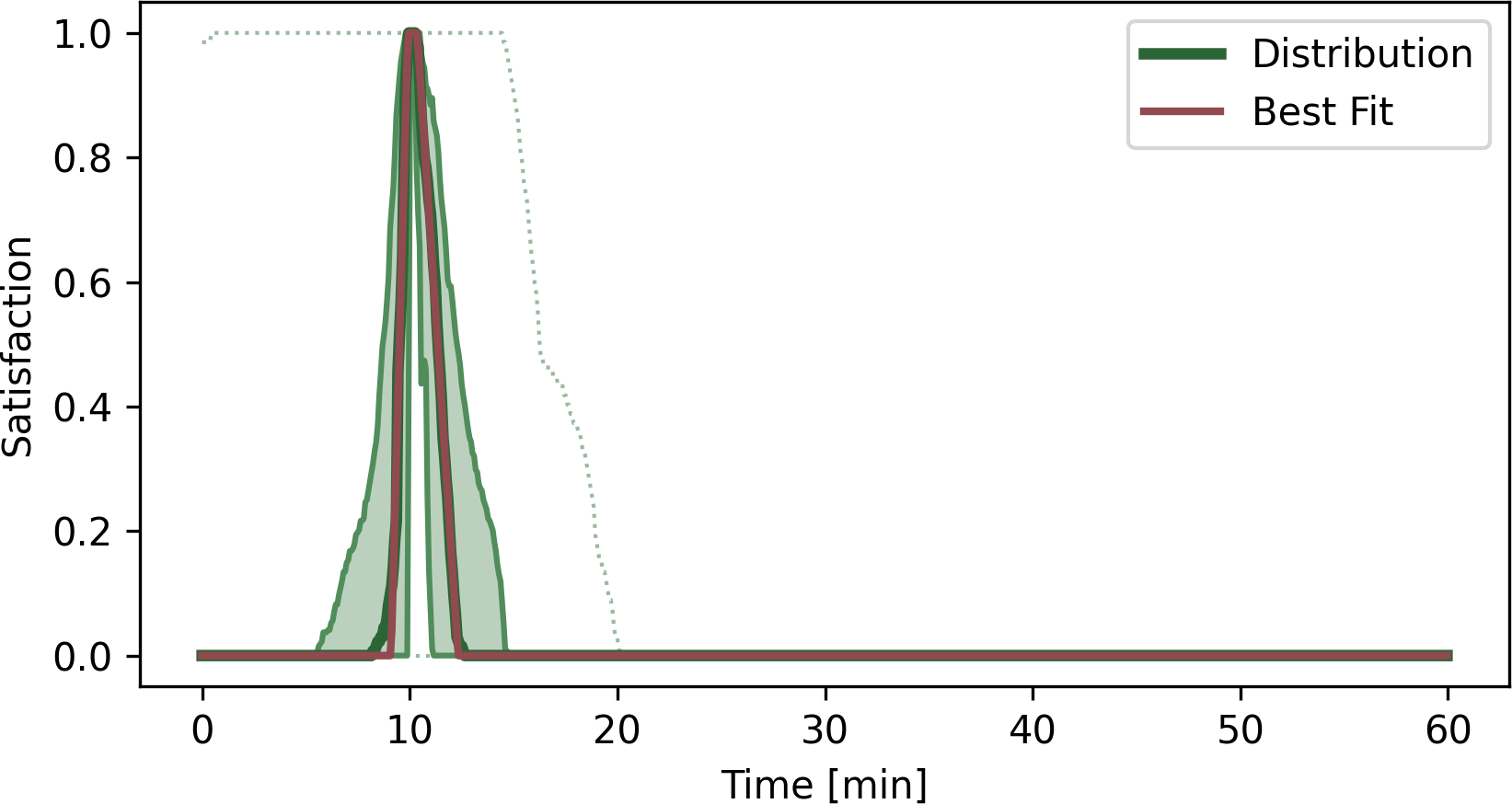}%
		\label{img:fit_in_ten}
	}
	\subfloat[``before ten minutes'']{
		\includegraphics[width=0.31\linewidth]{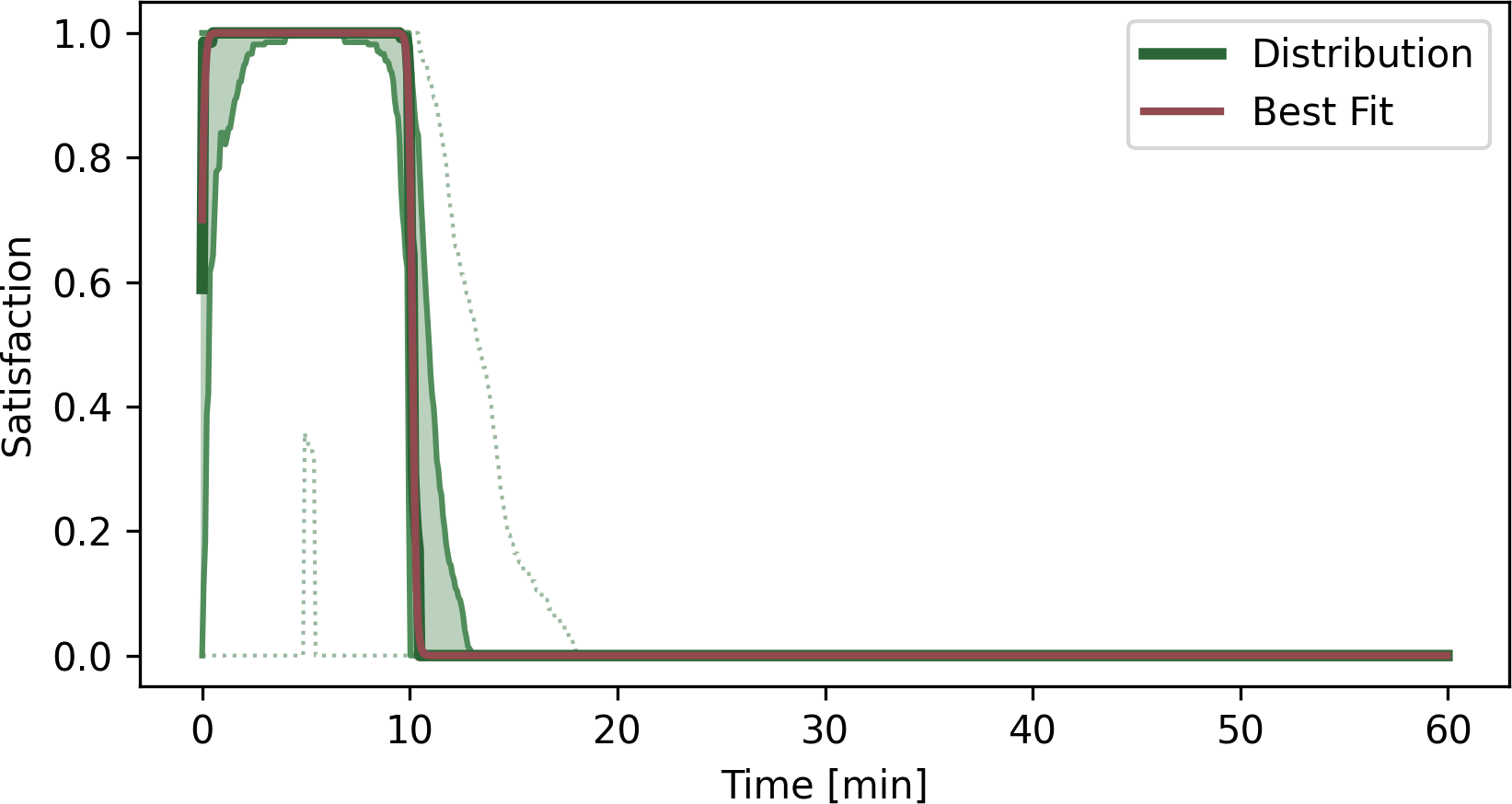}%
		\label{img:fit_before_ten}
	}
	\subfloat[``after ten minutes'']{
		\includegraphics[width=0.31\linewidth]{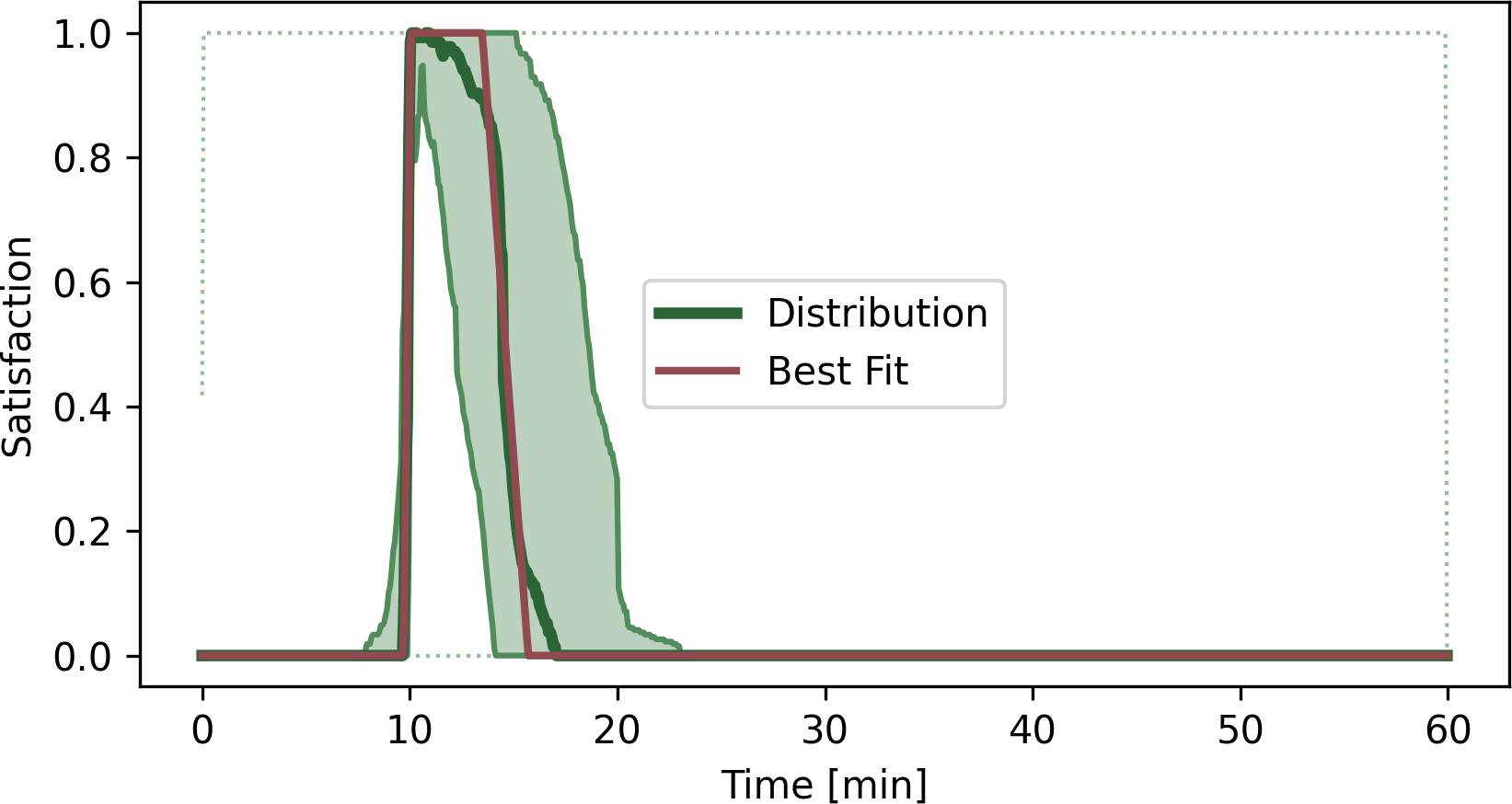}%
		\label{img:fit_after_ten}
	}
    \hfill
	\subfloat[``in 30 minutes'']{
		\includegraphics[width=0.31\linewidth]{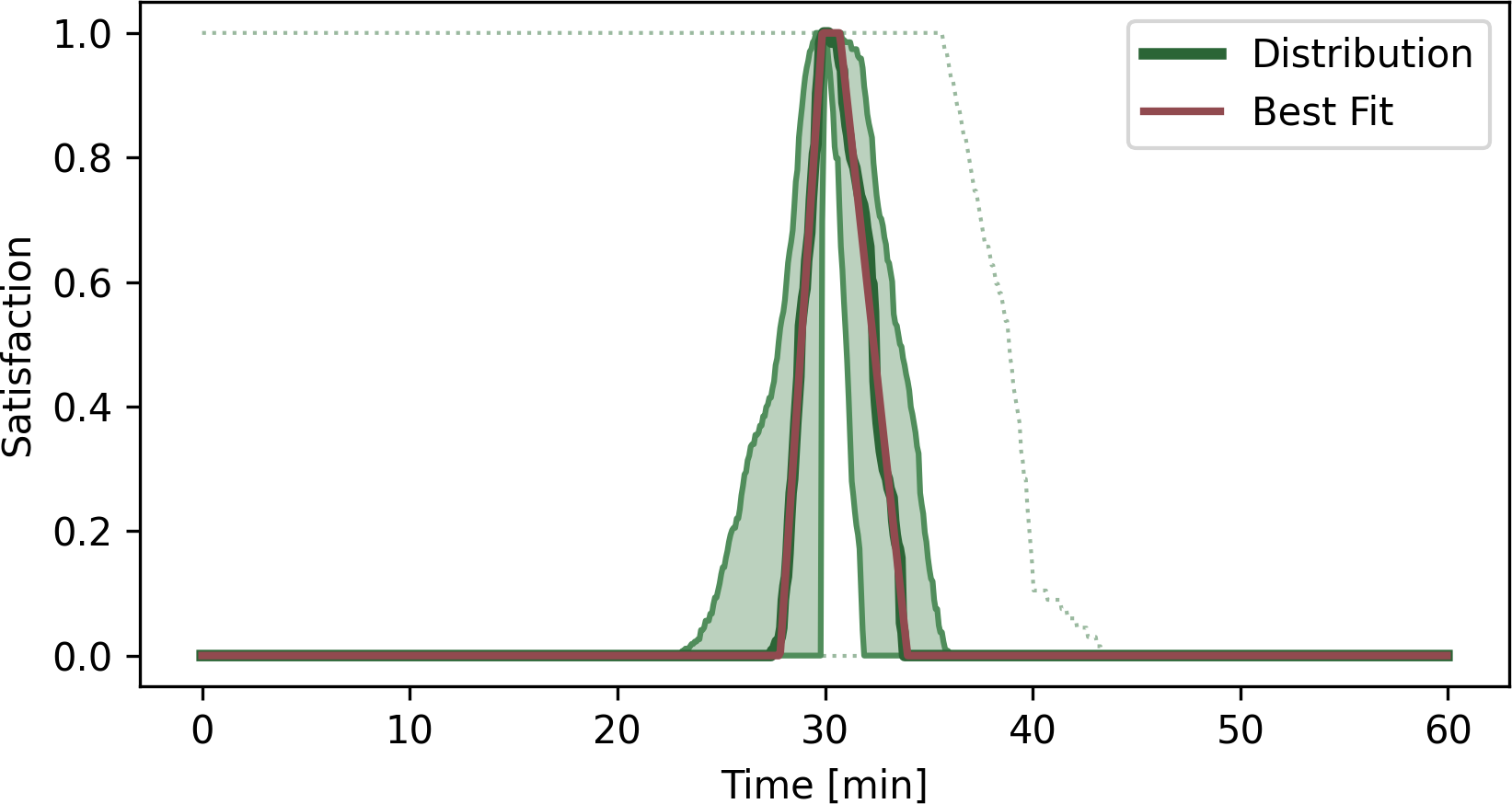}%
		\label{img:fit_in_thirty}
	}
	\subfloat[``before 30 minutes'']{
		\includegraphics[width=0.31\linewidth]{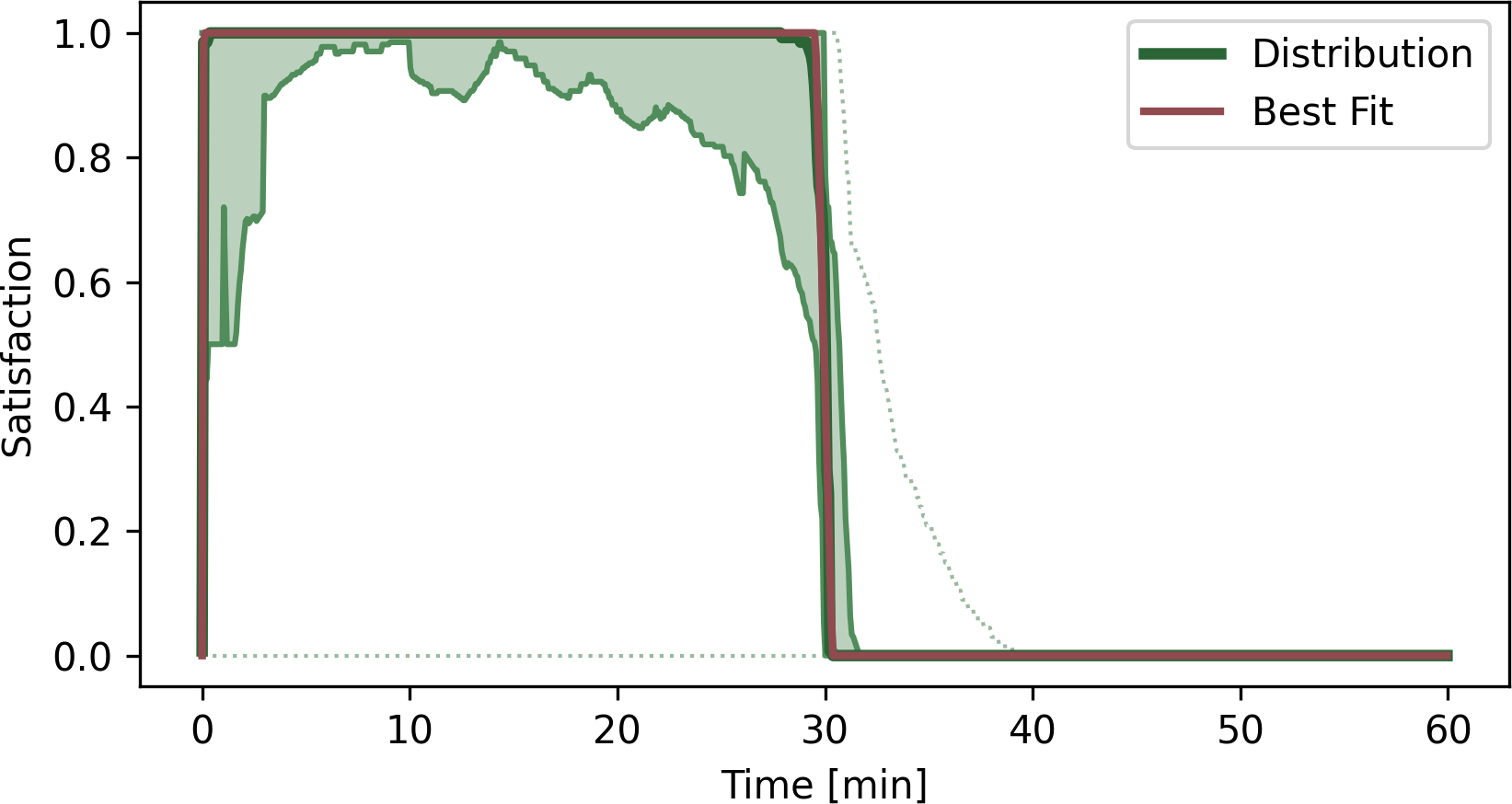}%
		\label{img:fit_before_thirty}
	}
	\subfloat[``after 30 minutes'']{
		\includegraphics[width=0.31\linewidth]{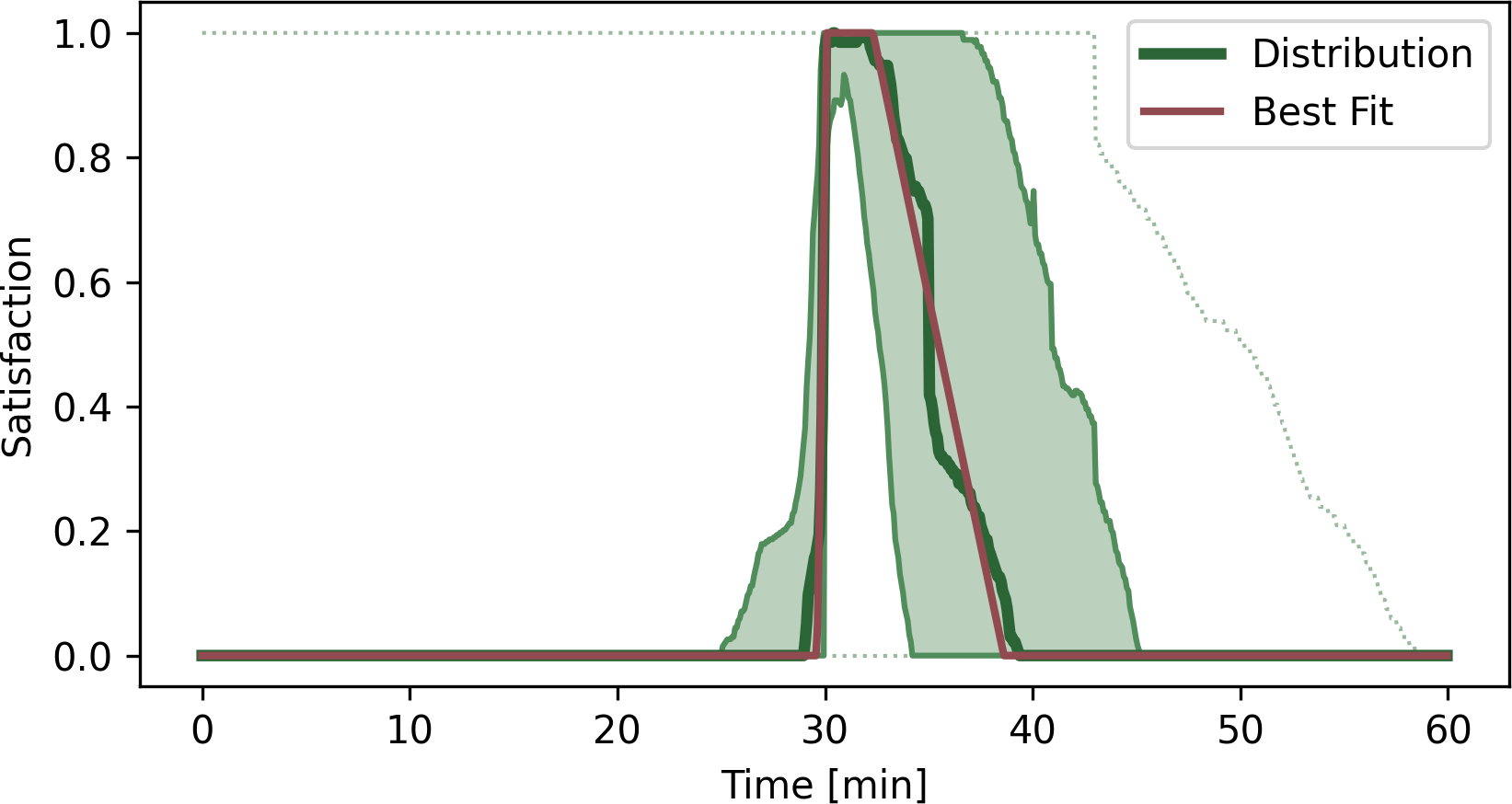}%
		\label{img:fit_after_thirty}
	}
	\caption{
		Excerpt of functions (red) fitted to the distributions (green). (a)-(c) represent the prepositions \textit{in}, \textit{before}, and \textit{after} at \textit{10 min}; (d)-(f) at \textit{30 min}. }
	\label{img:fitting}
\end{figure*}

\section{Evaluation} 
\label{sec:evaluation}
We investigate users' expectations of fuzzy time requirements by analyzing their satisfaction function given different instructions.
Corresponding to \autoref{sec:methodology:identifying}, we want to derive general satisfaction functions from subjective user data and approximate them with continuous functions.
Furthermore, we assume that various factors influence user satisfaction, such as the expected abilities of the actor or the instructed start time of the fuzzy time requirements.
This will be examined in detail to draw general conclusions about user satisfaction.
Based on this objective, we focus on three research questions:
\begin{itemize}
	\item[F1] Which continuous functions are best suited to model fuzzy time requirements with varying prepositions (in, after, before)?
	\item[F2] To what extent does the satisfaction function differ when the task is executed by a human compared to a robot?
	\item[F3] To what extent do different start times influence the satisfaction function?
\end{itemize}
Accordingly, we ask users to rate their satisfaction with the execution of instructions at various execution times.
A potential methodology would be a \textit{sampling-based} study design. 
This design requires users to instruct the actor, thus triggering the execution at a point in time. 
Then, the users would indicate their satisfaction with this execution time.
However, this has the disadvantage that we require many sample points (i.e., questions per instruction) for a high-resolution satisfaction function.
Assuming the satisfaction functions vary for different instructions, this design would require numerous questions.
Alternatively, a \textit{function-based} study design mitigates this issue:
Users receive an instruction and are asked to assess their own satisfaction over the entire timeline. 
The user then draws the complete satisfaction function instead of giving a few points to which the function is fitted.
This design may be less straightforward for the user as the execution is not directly shown.
However, this reduces the number of required questions, enabling a broader exploration of fuzzy time requirements, including a higher sampling rate and broader time interval.
We opted for a function-based study design as we could not collect sufficient data for the sampling-based design.

We conducted the so-designed study with an online user study.
The participants are directed that they are instructing one actor.
The task is to draw a satisfaction function over time for multiple instructions if the actor would start execution at the corresponding time.
To answer question F1, we compare their drawn satisfaction functions with fitted continuous functions.
We address question F2 by randomly assigning the participants into a 'robot' or 'person' actor group to avoid carryover effects.
Thus, the actor is consistently called either 'robot' or 'person' in the task description.
The instructions have different time specifications, allowing us to examine F3.

\subsection{User Study Design} 

The user study consists of four components: Demographic questions, an introduction, main tasks, and a control question.
Demographic questions include the age, gender, and cultural background of the participant.
The introduction explains the tasks and a short tutorial about the user interface (\autoref{img:tutorial_study}).  
In the described scenario, the actor is already working on other unnamed assignments. 
Therefore, a perfect adherence to the specified start times is not possible. 
The participants are each given 14 instructions and asked to draw their satisfaction if the actor would start the task at the corresponding time stamp.
The user study interface (\autoref{img:tutorial_study}) consists of the canvas for drawing and the timeline for quickly moving the canvas over time.
To mitigate the impact of the users' drawing skills with the computer mouse, they can overwrite the drawn function and, thus, precisely specify their desired function.

The instructions for the main tasks follow the pattern ``The assignment should start $\langle$preposition$\rangle$ $\langle$fuzziness$\rangle$ $\langle$time$\rangle$!'', where $\langle \cdot \rangle$ signifies placeholders: $\langle$preposition$\rangle\in$ \{\textit{in}, \textit{after}, \textit{before the next}\}, $\langle$fuzziness$\rangle\in$ \{$\emptyset$, \textit{approximately}\}, and $\langle$time$\rangle\in$ \{\textit{now}, \textit{one minute}, \textit{10 minutes}, \textit{30 minutes}\}.
Based on $\langle$time$\rangle$, every instruction contains the specified time $t_{\text{spec}}$ (e.g., \textit{one minute}: $t_{\text{spec}} = 1 \text{ min}$).
The order of the words in the instruction had to be partially adjusted for a correct grammatical sentence (e.g., ``The assignment should start before approximately the next ten minutes.'').
To reduce the testing fatigue of the participants while still investigating our research questions, we do not regard every combination of placeholders. 
Overall, we combined
\begin{inparaenum}[(i)]
    \item \textit{approximately} only with the times \textit{10-minutes} and \textit{now} (to \textit{soon}); and
    \item \textit{now} only with the \textit{in}-preposition. 
\end{inparaenum}
Thus, the user study consists of 14 instructions.
The timeline covers the time range between 0 and 60 minutes sampled by 800 measurements per instruction (sampling rate of $4.5$ $\text{s}^{-1}$).
The initial default value of the function at each time step is 0.
The instructions are presented to the participants in random order to minimize learning effects.
In advance, we defined as attention checks that the participant is filtered if the drawn function equals zero for every time step $\pm 5 \text{ min}$ around $t_\text{spec}$ in any instruction.
We deliberately provide generic instructions for result generalization, as the scenario may influence satisfaction.
The application scenario and explicit operations are intentionally kept generic. 
We assume that the application has a significant influence on user satisfaction.
Accordingly, we detached the tasks from a specific application to further generalize the results.
In the control question, the actor should start at 10 minutes but is three minutes late. 
The participants rate their satisfaction using a Likert Scale.

\subsection{Results and Discussion} 

The user study was conducted with convenience sampling, mostly among students and employees of the University of Bayreuth.
One person had to be filtered based on the attention check, resulting in 32 participants.
Of these, four are female (12.25\%), 26 are male (81.5\%), and two stated no answer (6.25\%). 
The participants are between 21 and 35 years old, with a median $\Tilde{x} = 26.5$ years.
Of all participants, 29 have a German cultural background (90.625\%), two have an Indian background (6.25\%), and one has a US-German background. 
The division into the actor group was 16 participants each.

The following sections are structured based on the research questions: 
Each section briefly explains the evaluation methodology, gives our predictions before the study commenced, presents the results, and discusses them.


\subsubsection{Best Fit Continuous Satisfaction Functions}
\label{sec:evaluation:best_fit}
We use a Trust Region Reflective (\autoref{sec:methodology:continuous_satisfaction}) to determine the best fitting functions.
Our study includes four instructions with \textit{in}-, \textit{after}-, and \textit{before}-preposition each.
We consider the respective tasks individually and compare the prepositions.
The functions are fitted to the median satisfaction function.
We considered rectangular, triangular, trapezoidal, and bell curve functions as they commonly model uncertainty in fuzzy-logic \cite{Zhao02}.
Since trapezoidal functions encompass rectangular and triangular functions, we fitted trapezoidal functions and bell curves.
We assumed that \textit{in}-prepositions lead predominantly to bell curves centered around $t_\text{spec}$.
We expected a trapezoidal to best approximate \textit{before}- (between 0 to $t_\text{spec}$ maximal) and \textit{after}-prepositions (between $t_\text{spec}$ and $t_\text{max}$ maximal).

The instructions with \textit{in} best fit to trapezoidal function with a prediction error between $0.084$ and $0.206$ (e.g., \autoref{img:fit_in_ten} and \autoref{img:fit_in_thirty}).
In contrast, the fitted bell curve's prediction error is between $0.221$ and $0.890$. 
The maxima of the trapezoid encompass $t_\text{spec}$. 
The fitted satisfaction functions to \textit{before}-instructions are divided into two bell curves (e.g., \autoref{img:fit_before_ten}) and trapezoidal functions (e.g., \autoref{img:fit_before_thirty}). 
Their errors are between $0.008$ and $0.098$.
The fitted parameters result in functions at maximum between $0$ and $t_\text{spec}$ (except for $t=0$ due to rounding error).  
Each fitted satisfaction function for \textit{after}-prepositions is trapezoidal with a prediction error between $0.038$ and $0.774$.
Analogous to \autoref{img:fit_after_ten} and \autoref{img:fit_after_thirty}, their right maximum is greater than $t_\text{spec}$, whereas the left maximum is directly on $t_\text{spec}$ (margin of $5.4$ s).
The mean of the absolute gradient on the left ($5.946$) is greater than on the right ($0.312$).


The participants' median is mostly linear, impeding the prediction quality of the bell curve fit.
Conforming with our expectations, the user satisfaction is maximum around $t_\text{spec}$.
We also confirmed the expectation regarding the \textit{before}-preposition as its value range between $0$ and $t_\text{spec}$ is maximal. 
Bell curves were also fitted, but their transition to $0$ is abrupt (similar to the trapezoidal functions).
Contrary to expectations, however, the median in \textit{after}-prepositions does not plateau from $t_\text{spec}$ to $t_\text{max}$.
Yet, it was represented with a trapezoidal function slowly decreasing to the right.
This indicates that participants are not satisfied to wait an extended period but have the execution close to $t_\text{spec}$.
The high difference between the quantiles of the participants' answers should be noted here (e.g., in \autoref{img:fit_after_thirty}), indicating divergent participant expectations.
In response to F1, trapezoids are suitable representations of fuzzy time requirements: 
\textit{in}-trapezoids are centered at $t_\text{spec}$; \textit{before}-trapezoids are maximal between 0 and $t_\text{spec}$, and \textit{after}-trapezoids are maximal at $t_\text{spec}$ decreasing slowly to the right.

\begin{figure}[t]
	\centering
	\includegraphics[width=0.96\linewidth]{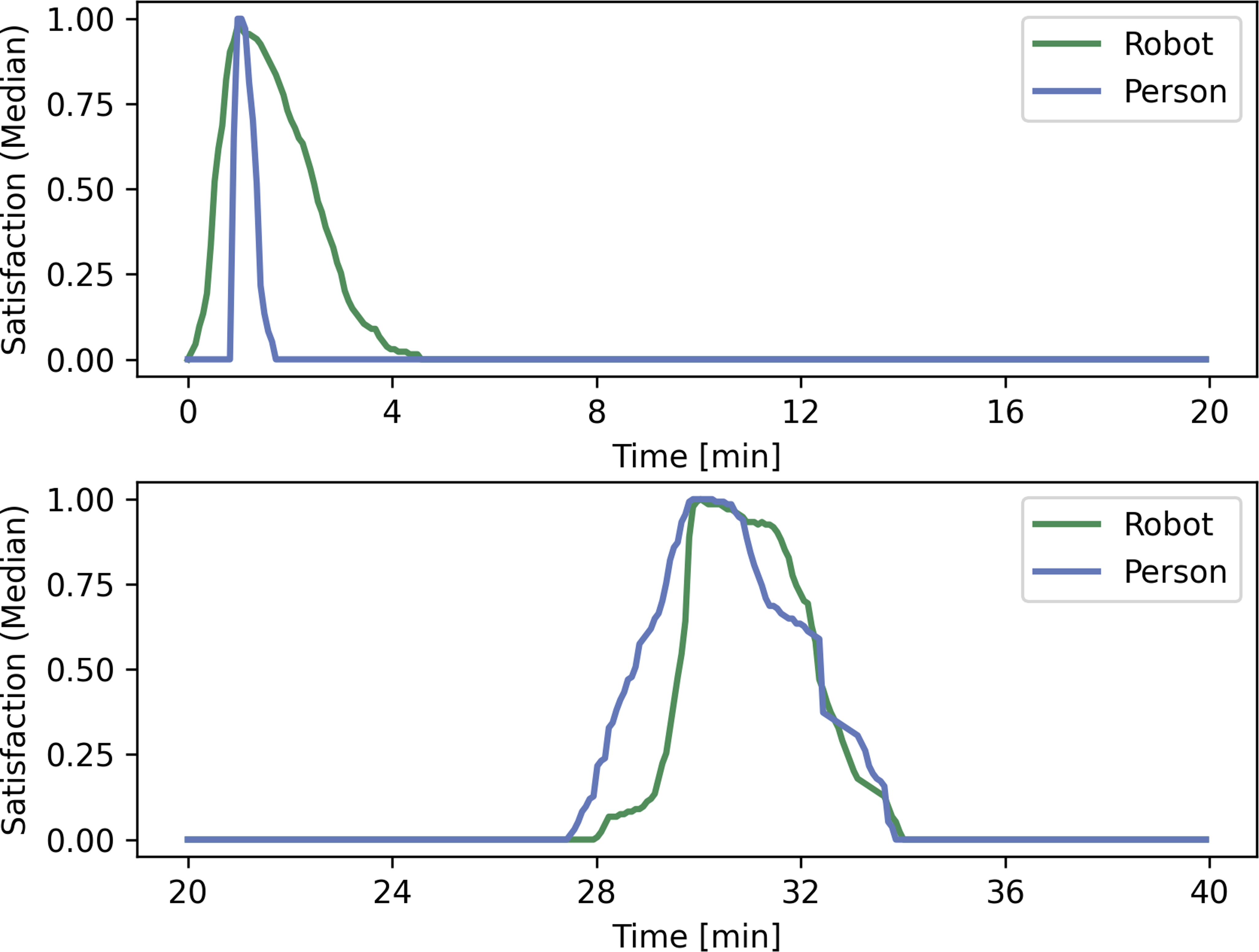}
	\caption{
		Comparison between robot and person actors. The time requirements are ``in one minute'' (top) and ``in 30 minutes'' (bottom). } 
	\label{img:robot_human_comparison}
\end{figure}

\subsubsection{Expectation Comparison between Human and Robot Actors}
\label{sec:evaluation:human_robot}
We compare the human and robot groups for each instruction. 
For this purpose, the median is calculated for each group independently.
We assume that the ``broadness'' of the respective satisfaction functions differs.
Accordingly, we want to calculate the variance of the satisfaction functions.
For this purpose, we interpret the median satisfaction functions as a probability density distribution over time, i.e., the percentage of satisfaction at one point in time.
This is compared between the groups.
Furthermore, the answers to the control question are compared and tested for significance with the Mann-Whitney U test.
We presumed that the participants are more lenient towards other people than towards a robot, as a robot is expected to be more precise.
This means that the variance of the person group is greater than that of the robot group.
We expect the participants to be significantly less satisfied with the robot's delays (control question).

The variance of the median satisfaction functions within the person group is smaller than in the robot group in 12 out of 14 cases.
The mean difference in variance between the person and robot group is $-0.769$ (negative values indicate a larger variance in the robot group).
In \autoref{img:robot_human_comparison} (top), the difference in variance for the \textit{in}-preposition is $-0.606$. 
An example of the robot group's lower variance with a difference of $0.633$ is shown in \autoref{img:robot_human_comparison} (bottom).
The control question's results exhibit a mode at $1$ (lowest) for the robot group and a mode at $3$ (medium) for the person group.
The Mann-Whitney U test results in a $p=0.267$.

The user study results cannot be unambiguously interpreted to which extent humans are more lenient towards persons than robots.
The robot group's variance is larger for most instructions, indicating greater leniency towards robots.
Humans perhaps consider other people as more capable than robots.
In contrast, the control question shows the participants' tendency to be more lenient towards people.
However, the test is not statistically significant ($p=0.267$ is over $0.05$).
Accordingly, we cannot answer question F2 unequivocally.
Nevertheless, we interpret the results as showing that the actor may influence user satisfaction.
Further studies with more actors (including different robot actors) could give more insight.
In particular, the differences between the direct opinion query (control question) and the drawn satisfaction functions must be evaluated.
It is also uncertain to what extent the absent visual component of the actor influenced our results.



\subsubsection{Influence of Start Times}
We compare the satisfaction functions of the four time-variants (\textit{now}, \textit{one minute}, \textit{10 minutes}, and \textit{30 minutes}).
We consider the \textit{in}-preposition without fuzziness modifier for a direct comparison of the start times.
For this, we plot the variance of the median over the time $t_\text{spec}$.
We presumed that the variance of the median and $t_\text{spec}$ are positively correlated, i.e., the greater the specified time, the greater the variance.
Punctuality may be less relevant for later instructions than for imminent ones.

\begin{figure}[t]
	\centering
	\includegraphics[width=\linewidth]{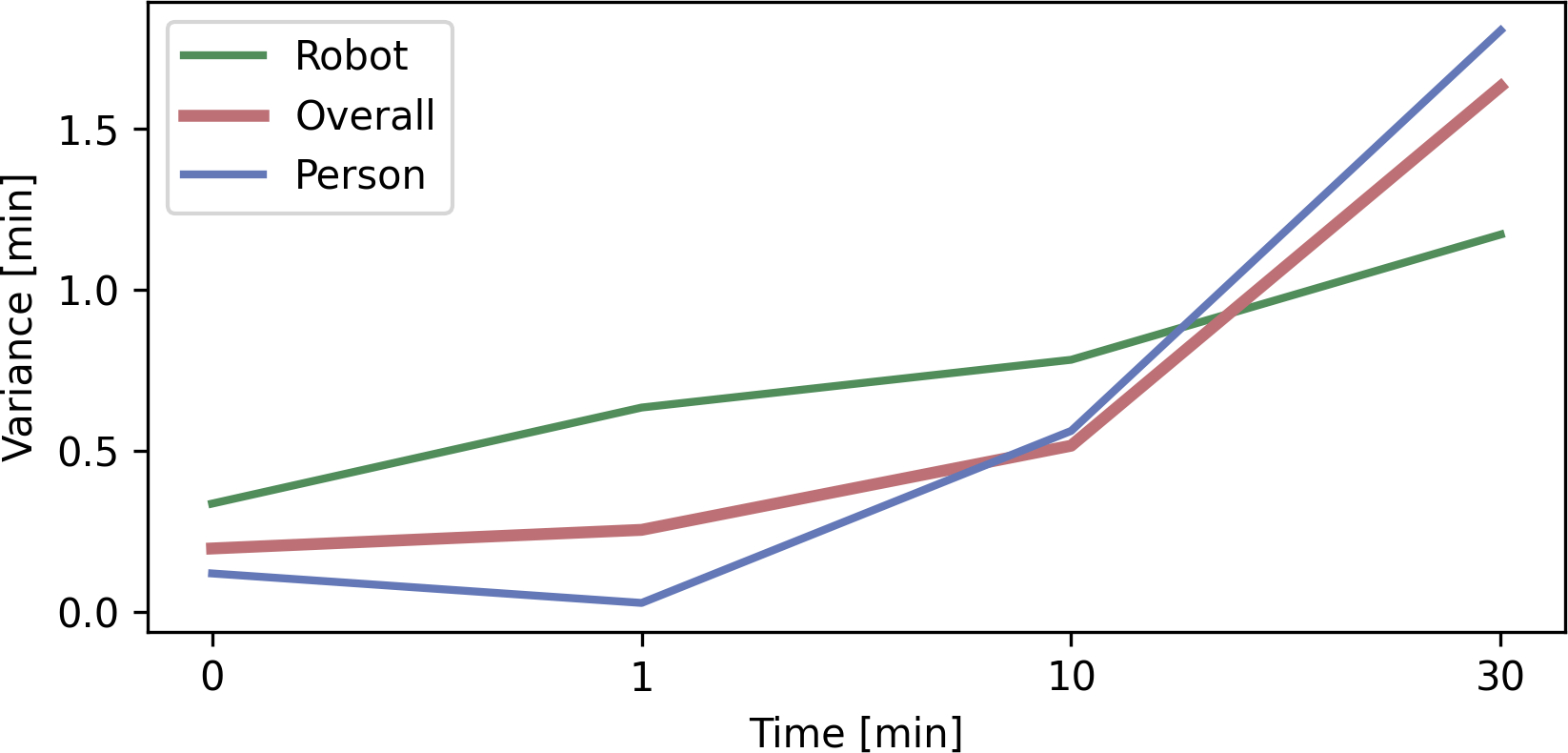}
	\caption{
		Comparison of how the specified start time influences the variance in the median satisfaction functions (at 0, 1, 10, and 30 minutes). } 
	\label{img:start_times}
\end{figure}

The variance across both groups increases with the specified time $t_\text{spec}$ (\autoref{img:start_times}).
For a \textit{now}-instruction, the general variance is 0.197; for \textit{in 30 minutes}, it is 1.63 (increased by a factor of $8.274$).
The robot variance at the markers 0, 1, and 10 minutes is higher, whereas, for 30 minutes, the person group's variance is 1.805 compared to 1.171.

The presumption that variance and $t_\text{spec}$ positively correlate is confirmed in the context of our evaluated time specifications.
Thus, users generally give the actor a larger execution window as it is specified further in the future.
For the person group, the variance is slightly smaller at $t_\text{spec}=1$ compared to $t_\text{spec}=0$. 
We attribute this result to noise.
Additionally, users have more tolerance for late executions by person actors in future operations (in \autoref{img:start_times} at 30 minutes).
Conceivably, users expect a more precise internal time in robots.
For this, further investigations with more time markers are required.

\section{Conclusions}
In this paper, we contribute a methodology for handling and interpreting fuzzy time requirements in natural language instructions.
We introduced fuzzy skills managing fuzziness in execution times (\autoref{sec:methodology:tafs}).
This fuzzy execution time is represented by satisfaction functions reflecting the user's satisfaction over time regarding the timeliness of execution.
We derive satisfaction functions from the user instruction by interpreting keywords and employing lookup methods or temporal fuzzy-logic (\autoref{sec:methodology:inferring}).
These functions enable the robot system to make educated decisions when specifications overlap, facilitating satisfaction-maximizing scheduling.
Hill Climbing approximates this optimization problem (\autoref{sec:methodology:scheduling}).
Satisfaction functions are subjective for the user. 
For generalization, we interpret subjective satisfaction functions by their characteristic properties and fit continuous functions to them (\autoref{sec:methodology:identifying}).
Building on this, we conducted an online user study (\autoref{sec:evaluation}) to investigate models for fuzzy time requirements, the influence of the actor (robot or person), and start time.
The participants assessed their satisfaction based on the actor's execution of a given instruction at different points in time.
For this, they would draw the complete satisfaction function, allowing a broad exploration of fuzzy time requirements.
The evaluation shows that trapezoidal functions are suitable for representing fuzzy time requirements. 
However, regarding the actor's influence, results are mixed: Users generally appear to grant the robot more leniency in execution time according to the drawn satisfaction functions. 
Even though the responses to the control question are not statistically significant, they contradict this, suggesting the opposite.
Start times further in the future increase the width of the tolerance window.

Our exploration of fuzzy time requirements lays the groundwork for future work:
We detached our evaluation from an explicit domain and hardware setup.
Accordingly, the conclusions must be validated based on an explicit scenario with real physical actors.
To gain further insight into the comparison between robot and human actors, a broader study with increased participant numbers and task variations can be conducted. 
Several aspects remain open for exploration, such as the impact of the application scenario, skill duration, indicator verbs (``finish the assignment by ...''), missing explicit time specification (e.g., ``soon''), and demographic factors (e.g., culture and education). 
Furthermore, the approaches to solve the optimization problem can be examined more closely, e.g., by analyzing their runtime and quality.
Additionally, a user study may investigate the automatic inference of the satisfaction functions from instructions in specific applications.

\section*{Acknowledgments}
The authors thank the participants of the user study for contributing to our research.

\addtolength{\textheight}{-2cm}   
\bibliography{IEEEabrv,references}

\end{document}